\documentclass{bmvc2k}
\usepackage{amsfonts}
\usepackage{amsmath}
\usepackage{amssymb} 
\usepackage{booktabs}
\usepackage{booktabs}
\usepackage{graphicx}
\usepackage{nicefrac}
\usepackage{pifont}
\usepackage{tabularx}
\usepackage{tikz}
\usepackage{url}
\usepackage{xcolor}
\usepackage{xspace}
\usepackage[export]{adjustbox}
\usepackage{hyperref}
\usepackage[capitalize]{cleveref}
\usepackage{wrapfig}
\usepackage{soul}
\hypersetup{colorlinks=true, urlcolor=magenta, citecolor=green}

\newif\ifdark
\IfFileExists{/Users/vedaldi/.bash_profile}{
\makeatletter
\immediate\write18{\unexpanded{%
if defaults read -g AppleInterfaceStyle 2>/dev/null;
then echo \\darktrue > /tmp/displaymode.tex;
else echo \\darkfalse > /tmp/displaymode.tex; fi}}
\makeatother 
\input{/tmp/displaymode.tex}
\ifdark
\definecolor{pcolor}{HTML}{1E1E1E}
\definecolor{tcolor}{HTML}{C5C5C5}
\else
\definecolor{pcolor}{HTML}{FDF6E3}
\definecolor{tcolor}{HTML}{333333}
\fi
\pagecolor{pcolor}
\color{tcolor}
\hbadness=\maxdimen
\vbadness=\maxdimen
\vfuzz=30pt
\hfuzz=30pt
}{}

\title{Guess What Moves: Unsupervised Video and Image Segmentation by Anticipating Motion}

\addauthor{Subhabrata Choudhury$^*$}{subha@robots.ox.ac.uk}{1}
\addauthor{Laurynas Karazija$^*$}{laurynas@robots.ox.ac.uk}{1}
\addauthor{Iro Laina}{iro@robots.ox.ac.uk}{1}
\addauthor{Andrea Vedaldi}{vedaldi@robots.ox.ac.uk}{1}
\addauthor{Christian Rupprecht}{chrisr@robots.ox.ac.uk}{1}

\addinstitution{
 Visual Geometry Group\\
 University of Oxford\\
 Oxford, UK
}
\runninghead{Choudhury$^*$, Karazija$^*$, Laina, Vedaldi, Rupprecht}{Guess What Moves}

\def\ie{\emph{i.e}\bmvaOneDot}
\def\eg{\emph{e.g}\bmvaOneDot}

\makeatletter
\newcommand{\cmark}{\ding{51}}%
\newcommand{\xmark}{\ding{55}}%
\makeatother

\newcommand{\DAVIS}{DAVIS\xspace}
\newcommand{\FBMS}{FBMS\xspace}
\newcommand{\ST}{STv2\xspace}

\makeatletter
\renewcommand{\paragraph}{%
  \@startsection{paragraph}{4}%
  {\z@}{0.5em}{-1em}%
  {\normalfont\normalsize\bfseries}%
}
\makeatother

\begin{document}
\maketitle

\def\thefootnote{*}\footnotetext{Authors contributed equally.}\def\thefootnote{\arabic{footnote}}

\begin{abstract}
Motion, measured via optical flow, provides a powerful cue to discover and learn objects in images and videos. However, compared to using appearance, it has some blind spots, such as the fact that objects become invisible if they do not move. In this work, we propose an approach that combines the strengths of motion-based and appearance-based segmentation. We propose to supervise an image segmentation network with the pretext task of predicting regions that are likely to contain simple motion patterns, and thus likely to correspond to objects. As the model only uses a single image as input, we can apply it in two settings: unsupervised video segmentation, and unsupervised image segmentation. 
We achieve state-of-the-art results for videos, and demonstrate the viability of our approach on still images containing novel objects.
Additionally we experiment with different motion models and optical flow backbones and find the method to be robust to these change. 
Project page and code available at \url{https://www.robots.ox.ac.uk/~vgg/research/gwm}.
\end{abstract} 
\section{Introduction}\label{s:introduction}
The motion of objects in a video can be detected by methods such as optical flow and used to discover and segment them.
A key benefit is that optical flow is object-agnostic: because it relies on low-level visual properties, it can extract a signal even before the objects are discovered, and can thus be used to establish an understanding of objectness.

The potential of motion as a cue is epitomized in \emph{video segmentation} problems, where the input is a generic video sequence and the task is to extract the main object(s) in the video.
In fact, some methods~\cite{yang2021self-supervised,meunier2022em-driven} adopt a motion-\emph{only} approach to video object segmentation, arguing that motion patterns are much easier to model and interpret than appearance. 
However, this approach ignores appearance cues and is `blind' to stationary objects.

Instead, we propose to use motion as \emph{supervision} to discover objects in videos \emph{and} still images without the need for manual annotations.
We observe that different objects tend to generate distinctive optical flow patterns which can be well approximated by small parametric models, such as affine or quadratic.
We use this fact to train a segmentation network that, given a \emph{single RGB frame} as input, predicts \emph{which image regions} are likely to contain such patterns.
The idea is that these regions would then separate the objects from the background.

This approach has several useful properties.
First, while motion is used for supervising the network, the latter implicitly learns the appearance of the objects, regularizing the segmentation.
Second, because the network works with a single image as input, it does not observe the motion directly. The model must anticipate what \emph{could} move, extracting objects even if they are not in motion.
Third, the network avoids predicting the objects' motion directly, which is a highly-ambiguous task given a single image as input; instead, it predicts only the support regions of the motion patterns, and the training loss measures the compatibility of such regions with the observed motion according to the assumed motion model.

While we are not the first to consider motion as a cue for decomposing an image into objects, our particular way of modeling motion is simple and versatile, and allows two application modes of our approach.
First, we consider \emph{internal learning for unsupervised motion segmentation}~\cite{ulyanov20deep}.
Given one or more videos as input (without labels), we optimize a network, as described above, to output a segmentation of the videos, effectively `observing' motion via backpropagation. %
Our approach achieves state-of-the-art performance on standard benchmarks for unsupervised motion segmentation~\cite{yang-loquercio2019unsupervised,yang2021self-supervised}.%

The second mode is \emph{transductive learning for unsupervised image segmentation}, which is intended to assess the generalization capabilities of our model as an image segmenter.
In this case, the network is first trained %
on a number of training videos and then evaluated on a disjoint set of images. %
Since only appearance information is available at test time, the problem solved is not motion segmentation, but image segmentation.
In this scenario, our model segments novel objects not observed during training, demonstrating the viability of our approach.

\section{Related Work}\label{s:related}

Our work aims to combine motion and appearance cues for unsupervised object discovery, in that motion can be used as a cue to learn a general object segmenter for both videos and images.
As such, there exist several related areas in literature, which we review next.

\paragraph{Unsupervised Video Object Segmentation.}
The aim of video object segmentation (VOS) is to densely label objects present in a video.
Current VOS benchmarks~\cite{perazzi2016a-benchmark-davis,li2013video-segtrack,ochs2014segmentation-fbms} usually define the problem as foreground-background separation, 
where the foreground comprises the most salient objects.
Efforts to reduce the amount of supervision follow two main directions, semi-supervised and unsupervised VOS.
Semi-supervised methods require manual annotations for the object(s) of interest in an initial frame during inference;
the goal is to re-localize these objects across the video~\cite{caelles2017one}.
Unsupervised VOS aims to discover object(s) of interest without the initial targets~\cite{faktor2014videonlc,papazoglou2013fast,tokmakov2019motion,jain2017fusionseg,li2018instance,lu2019see}.
However, 
most unsupervised VOS methods use, in fact, some form of supervised pre-training on external data. 

\paragraph{Motion Segmentation.}
In videos, the background is usually relatively static whereas objects in the scene have independent motion, thus providing a strong `objectness' signal.
Thus, many works approach unsupervised video object segmentation as a motion segmentation problem.
Several earlier methods address this problem by grouping point trajectories~\cite{brox10object,sundaram10dense,ochs2012higher,keuper2015motion,keuper2020motion,ochs2011object}, motion boundaries~\cite{papazoglou2013fast}, voting~\cite{faktor2014videonlc} and layered models~\cite{chang2013topology,jojic1993learning}.  %
More recently, \citet{Xie2022SegmentingMO, Lamdouar2021SegmentingIM} train motion models on generated scenes with synthetic 2D objects and generalize to real videos.
CIS~\cite{yang-loquercio2019unsupervised} proposes an adversarial framework, where an inpainter is tasked with predicting the optical flow of a segment based on context, while the generator aims to create segments with zero mutual information such that the context becomes uninformative.
DyStaB~\cite{yang2021dystab} extends CIS using the segmentation output of a dynamic model to bootstrap a static one.
In contrast to our method, this yields two separate models to choose from based on the application (\ie, video or static image segmentation). 
Instead, AMD~\cite{liu2021emergence} employs a single model with separate appearance and motion `pathways' and performs unsupervised test-time adaptation for video segmentation.   
Finally, MG~\cite{yang2021self-supervised} abandons the appearance pathway altogether, directly segmenting optical flow inputs with a Slot Attention-like architecture~\cite{locatello2020object}.
Closer to our approach, another line of work uses various motion models to group image regions.
Early methods~\cite{jepson1993mixture,torr1998geometric} consider mixture models of flow to account for the fact that a region may contain multiple motion patterns.
Another line of work~\cite{bideau2016s,bideau2018best} segments object translation directions from motion angle field obtained by correcting for estimated rotation of the camera. 
\citet{mahendran2018self-supervised} employ an affine flow model, using the entropy of flow magnitude histograms for loss to deal with noisy flow in real world. 
\citet{meunier2022em-driven} consider affine and quadratic motion models, however their method uses flows as input which makes it suitable only for videos during inference.

\paragraph{Unsupervised Image Segmentation.}
While we use motion as a learning signal, our method yields a general-purpose image segmentation network, separating an image into foreground and background, without using ground truth masks for supervision.
Early work in unsupervised image segmentation makes use of hand-crafted priors, \eg color contrast~\cite{cheng2015global,wei2012geodesic}, while some recent methods also combine handcrafted heuristics to generate pseudo-masks and use them to train using deep networks~\cite{zhang2018deep,zeng2019multi,nguyen2019DeepUSPS}.
Others address this problem via mutual information maximization between different views of the input~\cite{ji19invariant,Ouali_2020_ECCV}. 
A recently emerging line of work~\cite{bielski2019emergence,chen2019unsupervised,kanezaki2018unsupervised,benny2020onegan,voynov2021object,melas-kyriazi2022finding} explores generative models to obtain segmentation masks.
Many of them~\cite{bielski2019emergence,chen2019unsupervised,kanezaki2018unsupervised,benny2020onegan} are based on the idea of generating foreground and background as separate layers and combine them to obtain a real image.
Others~\cite{voynov2021object,melas-kyriazi2022finding} analyze large-scale unsupervised GANs (\eg BigGAN~\cite{brock2019large}) and find implicit foreground-background structure in them to generate a synthetic annotated training dataset.
Alternative line of work explores feature maps of self-supervised Vision Transformers, such as DINO~\cite{Caron_2021_ICCV}. 
For example, STEGO~\cite{hamilton2022unsupervised} supports segmenting multiple \emph{classes} in an image, performing \emph{semantic} segmentation, by distilling features and class centroids from DINO.
In \citet{melas-kyriazi2022deep} and TokenCut~\cite{wang2022selfsupervised}, authors model image patches with an affinity graph based on DINO feature alignment and perform further analysis on this graph to extract masks. \citet{shin2022unsupervised} cluster features of a variety of self-supervised backbones to produce candidate masks, using them to train a segmenter. 
Instead, our model is trained on video data using optical flow as a supervisory signal.
However, since it only requires a single image as input at test time, we show that our method is applicable to this task, providing an alternative approach to unsupervised object segmentation. 

\paragraph{Unsupervised Object Discovery.} 
While the above methods often aim to segment the most salient object(s) in an image, unsupervised multi-object segmentation explores the problem of decomposing a scene into parts, which typically include each individual foreground object and the background.
The usual approach is to learn structured object-centric representations, \ie to model the scene with latent variables (slots) operating on a common representation~\cite{greff2019multiobject,locatello2020object,emami2021Efficient,lin2020space,crawford2019spatially,burgess2019monet,engelcke2019genesis,engelcke2021genesis,jiang2020generative,singh2022illiterate}.
While these methods are image-based, extensions to video also exist~\cite{kosiorek2018sequential,Jiang2020SCALOR,crawford2020exploiting,kabra2021simone,zablotskaia2020unsupervised,besbinar2021self,min2021gatsbi,bear2020learning,kipf2021conditional,singh2022simple,monnier2021dtisprites}.
These methods often operate in an auto-encoding fashion with inductive bias to separate objects derived from a reconstruction bottleneck~\citep{burgess2019monet}, that is often dependent on the architecture and the latent variable model.
We similarly impose a reconstruction bottleneck on the flow 
but use a simple model grounded in projective geometry, with a known closed-form solution.
It is also important to note that unsupervised \emph{multi}-object segmentation appears to be significantly more challenging, with current methods exploiting the simplicity of synthetic scenes~\cite{johnson2017clevr,Girdhar2020CATER}, %
while struggling on more realistic data~\cite{karazija2021clevrtex}. 
Recently, \citet{bao2022discovering} explore an extension of Slot Attention~\cite{locatello2020object}, guided by an external supervised motion segmentation algorithm, to real-world data. 
However, due to the difficulty of the problem, they operate in a constrained domain (autonomous driving) and consider only a limited number of object categories. 
We instead focus on wide variety of categories and settings encountered in common video segmentation datasets and consider both motion and appearance jointly. 

\section{Method}\label{s:method}

\begin{figure}[t]
    \centering
    \includegraphics[width=0.95\textwidth,trim={2mm 45mm 27mm 1mm},clip]{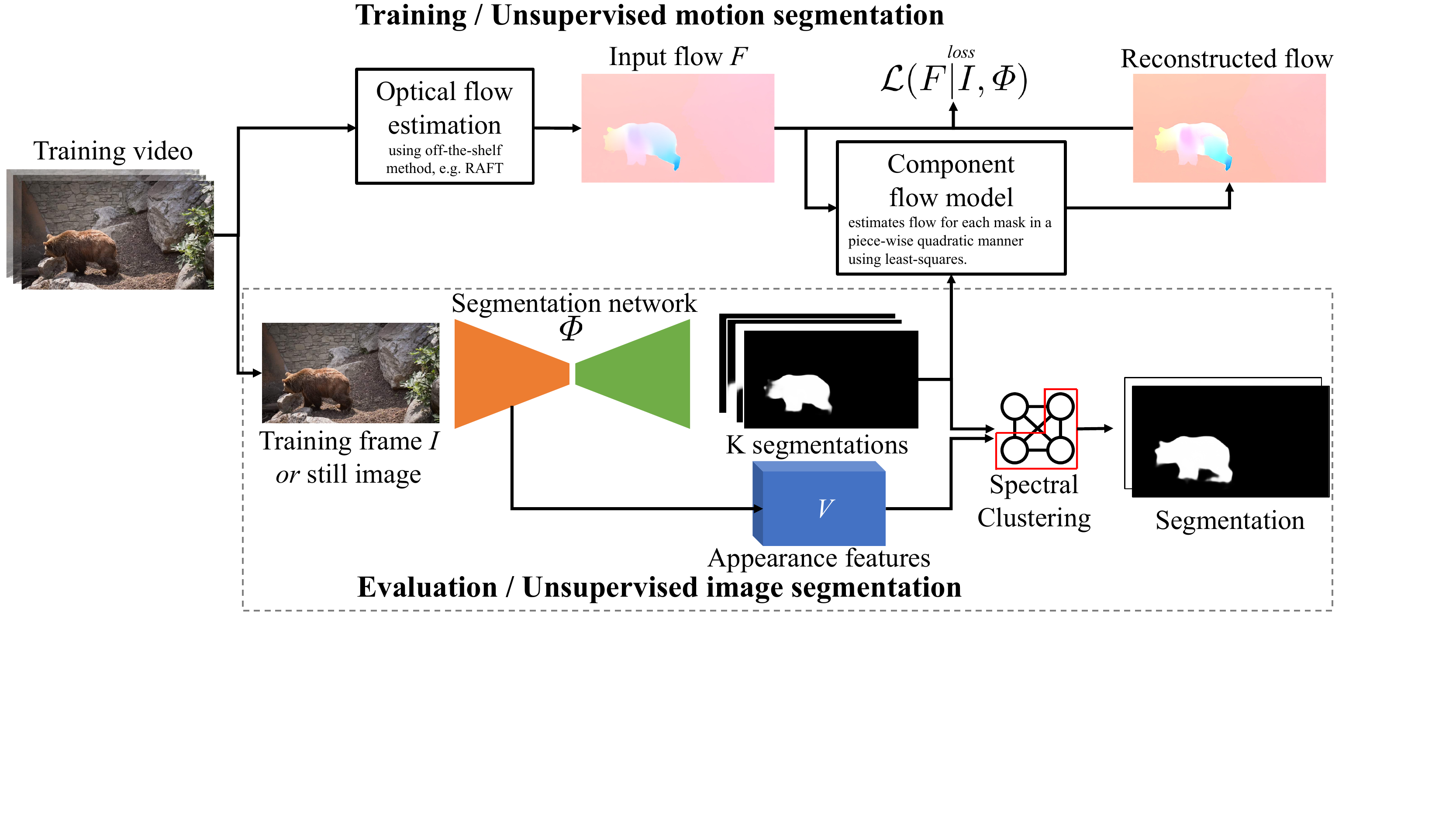}
    \vspace{-1em}
    \caption{\textbf{Model Diagram.} We train a segmentation network to partition an image into $K$ components without manual annotations. Our model is trained using individual frames from video as input and pre-computed optical flow as supervision. 
    The predicted segments are used to approximate the input flow with piecewise quadratic flow models and the training loss is formulated as the error between the reconstructed and the input flow.
    Appearance features from the backbone are used to merge the predicted $K$ segments into foreground and background components.
    Motion information is not required at test time and inference can be performed on still images. 
    Optical flow is colorized for visualization only. %
    }
    \label{fig:teaser}
\end{figure}

In this paper, we present a method that uses motion anticipation to discover and segment objects in images without the need for human annotations (overview in \cref{fig:teaser}). 
We use optical flow from video sequences as supervision for this problem.
However, rather than predicting the flow directly, we task a general image segmentation network to predict image regions where motion may be explained by a simple coherent model. Such regions should align with optical flow patterns produced by objects that \emph{could} move (but do not have to). 

\subsection{Segmentation by Motion Anticipation} %
Let %
$
I \in \mathbb{R}^{3\times H\times W} = (\mathbb{R}^3)^\Omega
$
be an RGB image defined on a lattice
$
\Omega = \{1,\dots,H\} \times \{1,\dots,W\}.
$
Assume that the image is a frame in a video sequence and let
$
F \in (\mathbb{R}^2)^\Omega
$
be the corresponding optical flow (extracted from the video by means of an off-the-shelf optical flow network, such as RAFT~\cite{teed2020raft}).
The goal is to decompose the image into $K$ components (or regions), which is a classic segmentation problem. 
Hence, we learn a segmentation network $\Phi(I) \in ([0, 1]^K)^\Omega$ that, given the image $I$ as input, assigns each pixel $u$ to one of $K$ components in a soft manner, with probabilities:
\begin{equation}\label{e:prob_output}
P(m_u=k \mid I,\Phi) = [(\Phi(I))_k]_u,
~~~
u \in \Omega, \; k \in \{1,...,K\}.
\end{equation}
$m_u=k$ in \cref{e:prob_output} denotes the predicted mask corresponding to component $k$ indexed by $u$.
In particular, we seek to separate the foreground and background, for which one may choose $K=2$, although as we show later (\cref{ss:overseg}), this need not be the case.

More specifically, we train $\Phi$ to partition pixels according to the Gestalt principle of common fate~\cite{spelke1990principles,wagemans2012century}.
This is done by associating each region $k \in \{1,...,K\}$ to a model $\theta_k$ of the optical flow observed within it. 
That is, the optical flow corresponding to an input frame can be approximated by piece-wise parametric models, representing the motion or \emph{flow pattern}, of each component independently.
According to the common fate principle, pixels within the same region are expected to exhibit \emph{coherent} motion.

A variety of motion models %
exist for describing the 2D flow of an object (\cite{mahendran2018self-supervised,meunier2022em-driven}). %
These are generally of the form $F_u \approx A u + b$, where parameters $A, b$ can be recovered by solving a system of linear equations. 
One common choice is an affine model (where $u = [x,y]$ are pixel coordinates), which is sufficient if objects are smaller and further away from camera. 
The affine model, however, struggles if the depth of an object varies significantly resulting in more complex flow patterns. 
To factor out unknown depth information, each object can be modeled as a plane with a quadratic 8-parameter model~\citep{adiv1985determining}.  
Here, we allow for more complex geometry than planes, by using a simplified 12-parameter quadratic model $\theta_k = (A_k, b_k)$ with $A_k \in \mathbb{R}^{2\times 5}$ and $b_k \in \mathbb{R}^2$ per region $k$. 
In this case, $u= [x,~x^2,~y,~y^2,~xy] \in \mathbb{R}^5$ includes quadratic and mixed terms of the pixel coordinates to model quadratic dependencies.
The 12-parameter model also allows treating each flow direction independently.
We assume that the model predicts the flow up to isotropic i.i.d.~Gaussian noise, which results in a simple $L^2$ fitting loss:
\begin{equation}\label{e:affine}
- \log p(F_u \mid \theta_k)
\propto
\|F_u - A_ku-b_k\|^2.
\end{equation}
Summing over all pixels, learning minimizes the energy function:
\begin{equation}\label{e:reconstruction_target}
  \mathcal{L}(F \mid \theta, I,\Phi)
  \propto
  \sum_{u \in \Omega}  \sum_{k}
  \|F_u - A_{k}u-b_{k}\|^2  \cdot p(m_u = k \mid I,\Phi) .
\end{equation}
In the expression above, we \emph{do not} know the flow parameters $\theta_k$ as the network only predicts the regions' extent.
Instead, we \emph{min-out} the parameters $\theta_k$ in the loss itself and compute
\begin{equation}\label{e:loss}
  \mathcal{L}(F\mid I,\Phi) = \min_{\theta_{k \in \{1,...,K\}}} \mathcal{L}(F \mid \theta, I,\Phi).
\end{equation}
The energy in \cref{e:affine} is quadratic in $\theta_k$, resulting in a weighted least squares problem that can be efficiently solved in closed form (see supplementary material).

Our model is learned from a large collection $\mathcal{T}$ of video frame-optical flow pairs $(I,F)$, minimizing the empirical risk:
\begin{equation}\label{e:final_loss}
\Phi^* =\operatornamewithlimits{argmin}_\Phi 
\frac{1}{|\mathcal{T}|}\sum_{(I,F)\in\mathcal{T}} 
\mathcal{L}(F \mid I,\Phi)
\end{equation}

\subsection{Over-segmentation}\label{ss:overseg}
While the 12-parameter model is more powerful than an affine one, it is still not sufficient to model arbitrary flow patterns.
In complex scenes that contain foreground and background clutter, we often observe motion parallax effects. 
Additionally, non-rigid objects and self-occlusions can result in complex flow patterns within the object that are not captured accurately by the quadratic model.

To account for such complexity, we propose to \emph{over-segment} the input image into $K > 2$ regions. 
Over-segmentation enables the model to use additional regions to explain several moving objects and to approximate varyingly moving parts of a single non-rigid object as well as motion parallax. 
To achieve a binary segmentation output, one needs a criterion to merge a number of predicted regions down to foreground and background.

We devise a criterion based on \emph{appearance} cues to avoid the ambiguity associated with merging regions based on motion.
To this end, we use a pre-trained self-supervised image encoder, such as DINO-ViT~\cite{Caron_2021_ICCV}, to obtain dense features for the input image and merge the segments predicted by $\Phi$ based on feature similarity.
Formally, let $V_u$ denote the feature vector of pixel $u$ obtained by the self-supervised encoder. 
Then, $\bar V_k = \sum_u V_u p(m_u=k \mid I,\Phi) / \sum_v p(m_v=k \mid I,\Phi)$ is the average feature vector for segment $k$, where pixels are weighed by their probability with which they belong to the segment. We compute the pairwise similarities of different regions via an affinity matrix $\Pi \in \mathbb{R}^{K \times K}$, where entries corresponding to segments $i$ and $j$ are set as 
\begin{equation}
    (\Pi)_{ij} = \max \left(\epsilon,  \left\langle \frac{\bar V_i}{||\bar V_i||_2},\frac{\bar V_j}{||\bar V_j||_2}\right\rangle \right),
\end{equation} 
where only feature vectors pointing in the same direction are considered and  $\epsilon = 10^{-12}$ is a small constant that keeps the graph connected. We then perform spectral clustering \cite{cheeger1969lower,shi2000normalized,melas-kyriazi2022deep} into two components using the affinity $\Pi$.

\subsection{Two Scenarios: Motion \emph{vs} Image Segmentation}
\label{ss:modes}

We experiment with two modes of application of our model. %
The first scenario is \emph{internal learning for unsupervised video segmentation}, where 
the network is evaluated on %
the same video sequences that have been used for optimization. %
This is effectively an unsupervised motion segmentation algorithm because the network not only receives as input appearance information, but incorporates motion information via backpropagation, observing indirectly optical flow too.
While not explicitly stated in the respective papers, prior motion segmentation works such as~\cite{yang2021self-supervised,meunier2022em-driven} also operate in this mode, while directly observing moving objects, often using optical flow as input.

The second scenario is \emph{transductive learning for image segmentation}.
In this case, the network is first trained using a number of unlabelled videos, and then used for single-image foreground object segmentation on an \emph{independent} validation/test set of still images.
In this scenario, motion is only used as a supervisory signal: when the network is applied at test time, motion is not considered anymore and the network operates purely as an image-based segmenter.
As for any transductive learning setting, the goal is to assess the generalization performance of the network on new images.

\section{Experiments}\label{s:experiments}

\begin{table*}[t]
\footnotesize
\begin{center}
\renewcommand*{\arraystretch}{0.9}
\begin{tabular}{@{}r@{\hspace{2pt}}l@{\hspace{20pt}}c@{\hspace{3pt}}c@{\hspace{3pt}}ccc@{\hspace{3pt}}ccc}
\toprule
                                               &                              & \multicolumn{2}{c}{Inf. Input} & Input & Flow & Runtime & \textbf{\DAVIS{}}   & \textbf{\ST{}}         & \textbf{\FBMS{}} \\
                                               &                              & RGB                       & Flow    & Resolution            & Method &  sec $\downarrow$   & $\mathcal{J}\uparrow$ & $\mathcal{J}\uparrow$ & $\mathcal{J}\uparrow$ \\
\midrule
{}\cite{sage}                                  & SAGE                         & \cmark                    & \cmark             & -- & LDOF~\cite{brox2010ldof} & $0.9$        & 42.6                  & 57.6                  & 61.2 \\
{}\cite{faktor2014videonlc}                    & NLC                          & \cmark                    & \cmark             & -- & SIFTFlow~\cite{liu2010sift} & $11$       & 55.1                  & 67.2                  & 51.5 \\
{}\cite{keuper2015motion}                      & CUT                          & \cmark                    & \cmark             & -- & LDOF~\cite{brox2010ldof} & $103$        & 55.2                  & 54.3                  & 57.2 \\
{}\cite{papazoglou2013fast}                    & FTS                          & \cmark                    & \cmark             & -- & LDOF~\cite{brox2010ldof} & $0.5$                 & 55.8                  & 47.8                  & 47.7 \\
{}\cite{yang-loquercio2019unsupervised}        & CIS                          & \cmark                    & \cmark             & $192 \times 384$ & PWCNet~\cite{sun2018pwcnet} & $0.1$                & 59.2                  & 45.6                  & 36.8 \\
{}\cite{liu2021emergence} & AMD & \cmark & \xmark & $128 \times 224$ & \xmark & -- &  57.8 & 57.0 & 47.5 \\ 
{}\cite{yang2021self-supervised}               & MG                           & \xmark                    & \cmark             & $128 \times 224$ & RAFT~\cite{teed2020raft} & $0.012$                 & 68.3                  & 58.6                  & 53.1 \\
{}\cite{meunier2022em-driven}                  & EM                           & \xmark                    & \cmark             & $128 \times 224$ & RAFT~\cite{teed2020raft} & --                 & 69.3                  & 55.5                  & 57.8 \\
{}\cite{Xie2022SegmentingMO}                  & OCLR                           & \cmark                    & \cmark             & $480 \times 832$ & RAFT~\cite{teed2020raft} & --                 & 78.9                  & 71.6                  & 68.7 \\
{}\cite{ye2022deformable}                  & {DS${}^\ddagger$} & \cmark                    & \cmark             & $240 \times 426$ & RAFT~\cite{teed2020raft} & $1800~(22.5)^\ddagger$                 & 79.1                  & 72.1                  & 71.8

\\[0.1cm]
\midrule
\multicolumn{2}{l}{\textbf{Ours} (UNet)}       & \cmark                       & \xmark                    & $128 \times 224$ & RAFT~\cite{teed2020raft} & $0.027$     & {78.3}     & {76.8}        & {72.0} \\
\multicolumn{2}{l}{\textbf{Ours} (MaskFormer)} & \cmark                       & \xmark                    & $128 \times 224$ & RAFT~\cite{teed2020raft} & $0.059$     & {\bf 79.5} & {\bf 78.3}    & {\bf 77.4} \\
\midrule
\midrule
{}\cite{yang-loquercio2019unsupervised}     & {CIS${}^\dagger$}     & {\cmark}     & {\cmark}  & $192 \times 384$ & PWCNet~\cite{sun2018pwcnet} & $11$     & {71.5}  & {62.0}  & {63.5} \\
{}\cite{yang2021dystab}     & {DyStaB${}^\dagger$*}     & {\cmark}     & {\cmark}  & $192 \times 384$ & RAFT~\cite{teed2020raft} & --     & {80.0}  & {74.2}  & {73.2} \\
\midrule
\multicolumn{2}{l}{{\textbf{Ours${}^\dagger$} (w/ CRF)}} & {\cmark}   & {\xmark}  & $128 \times 224$ & RAFT~\cite{teed2020raft} & $3.73$ &  {\bf 80.7} & {\bf 78.9} & {\bf 78.4}
\\
\bottomrule
\end{tabular}%

\end{center}
\caption{
\small \textbf{Unsupervised video segmentation} on DAVIS2016, SegTrack-v2 (\textit{STv2}), and FBMS59. \\
$\dagger$ denotes the usage of CRFs and other extra significant post-processing (\eg, multi-step flow, multi-crop, temporal smoothing for CIS~\cite{yang-loquercio2019unsupervised}). $\ddagger$ DS is optimized per sequence; authors report 30 min training time for 80-frame video. * DyStaB utilises supervised pre-training. 
}
\label{tab:main}
\end{table*}

As discussed above, our formulation allows us to evaluate our method in two settings: video object segmentation and general image/object segmentation.
We show that learning a network that \textit{guesses what moves} not only results in state-of-the-art performance in video segmentation, but also generalizes to image segmentation without further training. %

\subsection{Experimental Setup}

\paragraph{Architecture.}

Our formulation enables us to use any standard image segmentation architecture for the model $\Phi$.
This has two main benefits: while training the model needs optical flow (and thus video data), inference can be performed on single images alone just like any image segmentation method.
Second, using a standard architecture allows us to benefit from (self-){}supervised pretraining, ensuring better convergence and broader generalization.
We experiment with both convolutional and transformer-based architectures.%

\paragraph{Datasets.}

For the \emph{video segmentation} task, we use three popular datasets: DAVIS2016 (\DAVIS)~\cite{perazzi2016a-benchmark-davis}, SegTrackV2 (\ST)~\cite{li2013video-segtrack}, as well as \FBMS~\cite{ochs2014segmentation-fbms}.
For the \emph{image segmentation} task, we consider the Caltech-UCSD Birds-200 (CUB) dataset~\cite{welinder2010caltech-ucsd-cub-200} and three saliency detection benchmarks: DUTS~\cite{wang2017learning-duts}, ECSSD~\cite{shi2016hierarchical-ecssd}, and DUT-OMRON~\cite{yang2013saliency-dut-omron}.

\begin{figure}[t]
    \centering
    \includegraphics[width=.95\textwidth]{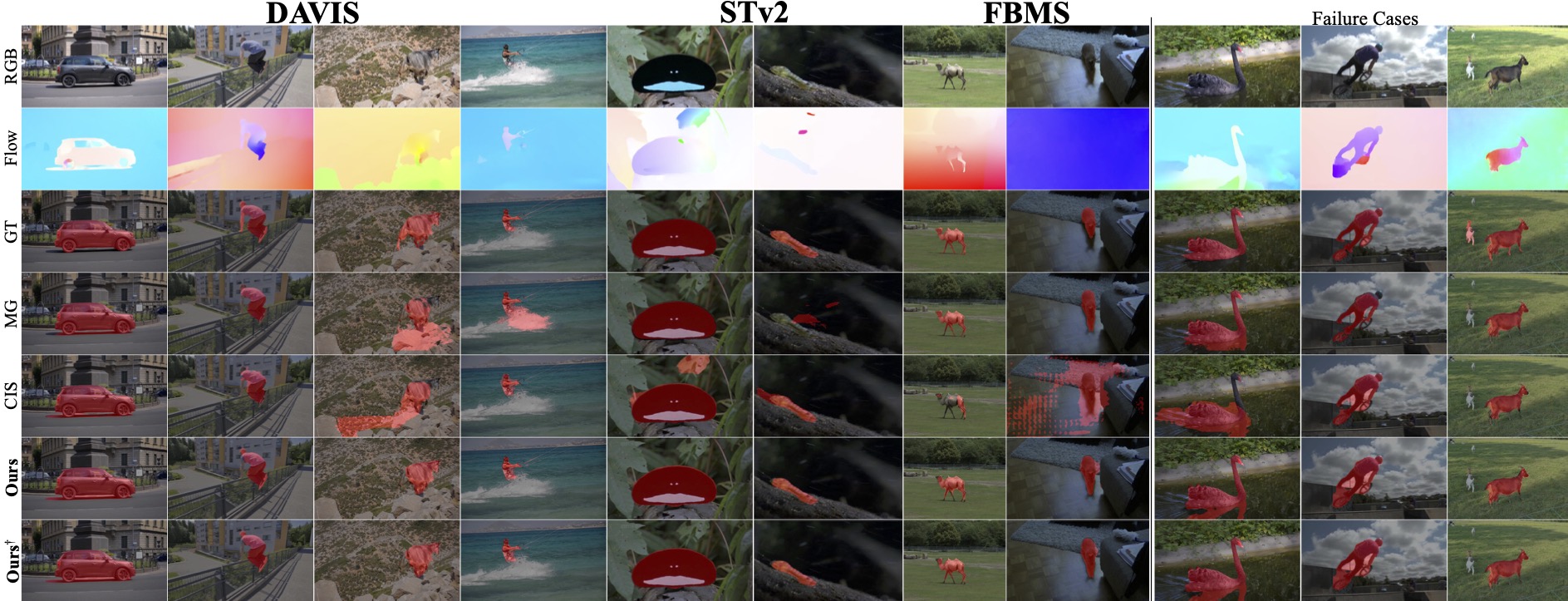}
    \caption{\textbf{Qualitative Comparison on \DAVIS, \ST, and \FBMS.}
    ${}^\dagger$-- indicates use of CRF. Our method correctly segments objects in challenging conditions including strong parallax (\textit{2${}^{nd}$, 3${}^{rd}$\@seq.}), small objects (\textit{4${}^{th}$}), background motion (\textit{5${}^{th}$}), camouflaged appearance (\textit{6${}^{th}$}), non-rigid motion (\textit{7${}^{th}$}) or no motion at all (\textit{8${}^{th}$} seq.). 
    In the failure cases, our method is confused by ripples and reflection in the water, the front wheel rotating in a different direction and multiple disconnected objects.}
    \label{fig:method_comparison}
\end{figure}

\paragraph{Optical Flow.}

Our method derives its learning signal from optical flow.
We estimate optical flow for all frames on \DAVIS, \ST, and \FBMS following the practice of MotionGrouping~\cite{yang2021self-supervised}.
We employ RAFT~\cite{teed2020raft} (supervised) using the original resolution for our main experiments. Please see the supplement for experiments with other flow methods.  

\paragraph{Training Details.}

We use MaskFormer~\cite{cheng2021maskformer} as our segmentation network, and use only the segmentation head.
For the backbone and appearance features $V$, we leverage a ViT-B transformer, 
pre-trained on ImageNet~\cite{russakovsky2015imagenet} in a self-supervised manner using DINO~\cite{Caron_2021_ICCV}
to avoid any external sources of supervision. We set the number of components to $K=4$ unless otherwise noted. Please see the supplement for all details and hyper-parameter settings.

\subsection{Unsupervised Video Segmentation}

In \Cref{tab:main} we report our performance on the \DAVIS, \ST, and \FBMS datasets and compare to other unsupervised video segmentation approaches.
Our method achieves state-of-the-art performance, even without CRF post-processing.
\cref{fig:method_comparison} provides a qualitative comparison of the results.
Our model provides better segmentation with sharper boundaries despite complex non-rigid motion, parallax effects or lack-of-motion.
However, on challenging scenarios our method still struggles to segment small details or non-connected instances.

Our method is not restricted to a specific segmentation architecture. To investigate, MaskFormer is replaced with a simple convolutional U-Net architecture~\cite{ronneberger2015u}, as in EM~\cite{meunier2022em-driven}, and trained from scratch for a fair comparison. The U-Net based model achieves comparable results on DAVIS and FBMS and 76.8 on STv2 (\cref{tab:main}), outperforming earlier methods even without transformers.

\subsection{Flow Model and Number of Components}\label{s:e:ablation}

\begin{table}[t]
\begin{center}
\footnotesize

\begin{tabular}{lccc}
\toprule
Flow Model & $K$ & \textbf{\DAVIS}~($\mathcal{J}$$\uparrow$) & \textbf{\DAVIS}~($\mathcal{J}_{\text{oracle}}$$\uparrow$) \\
\midrule
Constant ($A=0$)  & 4 & 76.8 & 77.7 \\ 
Affine ($u=[x,~y]$)  & 4 & 77.1 & 78.8 \\
Quadratic (\cref{e:affine}) & 4 & \textbf{79.5} & \textbf{81.5}\\
\midrule
Quadratic (\cref{e:affine}) & 2 & 74.5 & 74.5 \\
Quadratic (\cref{e:affine}) & 3 & 77.8 & 79.5 \\
Quadratic (\cref{e:affine}) & 4 & \textbf{79.5} & \textbf{81.5} \\
Quadratic (\cref{e:affine}) & 5 & 76.0 & 79.9 \\
\bottomrule
\end{tabular}%
\end{center}
\caption{\textbf{Flow Model and Number of Components.}
We ablate the choice of flow model and the number of components $K$. More complex flow models improve performance, and over-segmentation helps until the assignment problem between components and the final binary segmentation becomes too difficult at $K=5$. To evaluate the quality of the clustering of components we also report the oracle clustering performance as an upper bound.
}
\label{tab:loss_ablation}
\end{table}
Using \DAVIS, we now study the effectiveness of the individual components of the method.
In \cref{tab:loss_ablation} we evaluate the performance of the model under different flow models: constant, affine, and quadratic. 
We find that more complex models lead to improved performance, likely due to the fact that manny scenes in the \DAVIS benchmark are highly dynamic with complex objects and backgrounds.
Additionally, in the same table we evaluate how the number of components, $K$, influences the final performance after clustering. With $K=2$ the model directly performs foreground-background separation but needs to model each with a single component which is often difficult, \eg due to complex motions of deformable objects and/or parallax effects. Increasing the number of components is beneficial up to $K=4$, after which the assignment problem from over-segmentation to foreground and background becomes too difficult for simple spectral clustering.
This can be seen by evaluating the segmentation performance under an optimal oracle assignment of the components to foreground and background (oracle column in \cref{tab:loss_ablation}).
In all cases $K<=4$, spectral clustering nearly reaches oracle performance.

\begin{figure}[t]
\centering
\subfigure{
\begin{tabular}{p{6pt}c}
    & \scriptsize \textbf{CUB} \\[-1pt]
    \rotatebox[origin=c]{90}{\scriptsize Ours \hspace{25pt} GT  \hspace{25pt} Image } 
     & \includegraphics[page=1,height=0.19\textheight,valign=c]{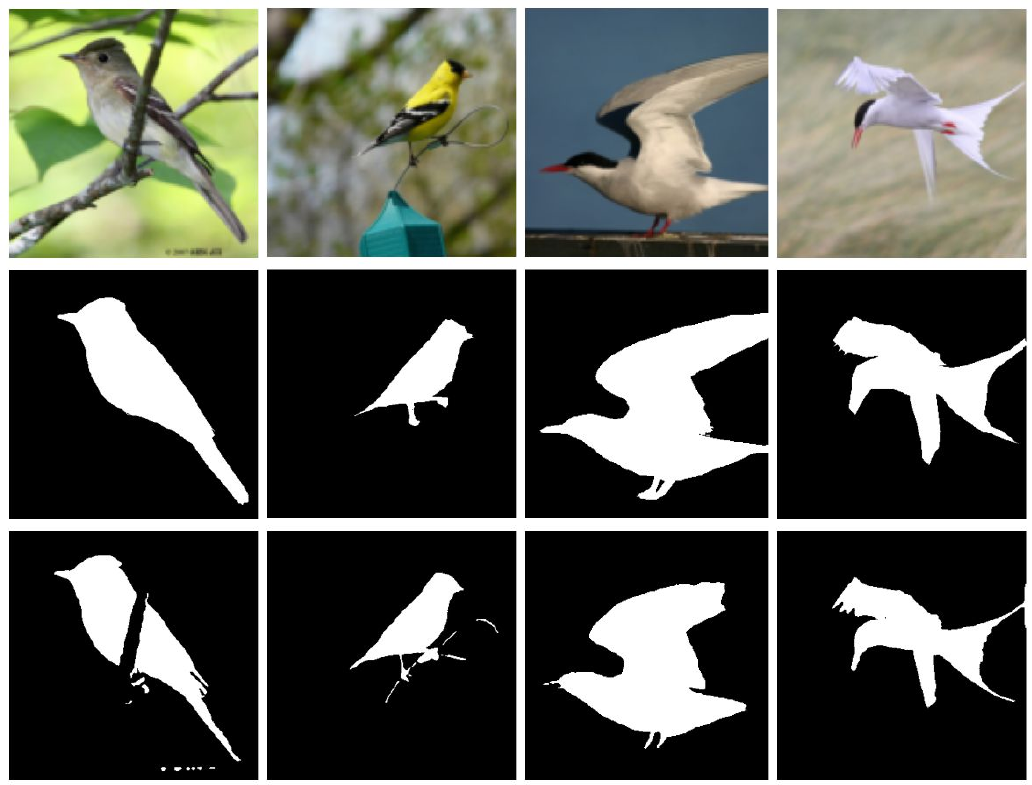} \\
\end{tabular}
}%
\subfigure{
\begin{tabular}{p{6pt}c}
    & \scriptsize \textbf{DUT-OMRON} \\[-1pt]
     \rotatebox[origin=c]{90}{\scriptsize Ours \hspace{25pt} GT  \hspace{25pt} Image} 
     & \includegraphics[page=2,height=0.19\textheight,valign=c]{figures/images/unsup-semseg-bmvc.pdf} \\
\end{tabular}
}
\\[-4pt]
\subfigure{
\begin{tabular}{p{6pt}c}
     & \scriptsize \textbf{DUTS} \\[-1pt]
     \rotatebox[origin=c]{90}{\scriptsize Ours \hspace{25pt} GT  \hspace{25pt} Image} 
     & \includegraphics[page=3,height=0.19\textheight,valign=c]{figures/images/unsup-semseg-bmvc.pdf} \\
\end{tabular}
}%
\subfigure{
\begin{tabular}{p{6pt}c}
    & \scriptsize \textbf{ECSSD} \\ [-1pt]
     \rotatebox[origin=c]{90}{\scriptsize Ours \hspace{25pt} GT  \hspace{25pt} Image} 
     & \includegraphics[page=4,height=0.19\textheight,valign=c]{figures/images/unsup-semseg-bmvc.pdf} \\
\end{tabular}%
}%
\caption{\textbf{Qualitative Comparison}. Our method can extract salient object in various environments and works even for novel object that were not included in the training data.  }
\label{fig:unsup_seg}
\end{figure}

\begin{table}[ht]
\footnotesize
\begin{center}
\setlength{\tabcolsep}{4.5pt}
\renewcommand*{\arraystretch}{0.95}
\begin{tabular}{@{}r@{\hspace{2pt}}lccc@{\hspace{9pt}}ccc@{\hspace{9pt}}ccc@{\hspace{9pt}}ccc} 
\toprule
& \multicolumn{1}{c}{}  & \multicolumn{3}{c}{\hspace{-20pt}\textbf{CUB}} & \multicolumn{3}{c}{\hspace{-20pt}\textbf{DUTS}} & \multicolumn{3}{c}{\hspace{-20pt}\textbf{ECSSD}}& \multicolumn{3}{c}{\hspace{-3pt}\textbf{OMRON}} \\ 
\cmidrule(r{10pt}){3-5}
\cmidrule(r{10pt}){6-8}
\cmidrule(r{10pt}){9-11}
\cmidrule{12-14}
                                    &                                                         & Acc           & $\mathcal{J}\uparrow$ & $max F_{\beta}\uparrow$  & Acc           & $\mathcal{J}\uparrow$  & $F_{\beta}\uparrow$  & Acc           & $\mathcal{J}\uparrow$  & $F_{\beta}\uparrow$   &    Acc           &$\mathcal{J}\uparrow$ & $F_{\beta}\uparrow$  \\
\midrule
{}\cite{voynov2021object}         & Voynov \emph{et~al.}                                                 & 94.0          & 71.0    & 80.7              & 88.1          & 51.1     & 60.0             & 90.6          & 68.4      & 79.0            & 86.0          & 46.4 & 53.3\\
{}\cite{liu2021emergence} & AMD & -- & -- & -- & -- & -- & 60.2 & -- & -- & -- & -- & -- & -- \\
{}\cite{melas-kyriazi2022finding}   & Kyriazi \emph{et~al.}                                            & 92.1          & 66.4      & 78.3            & 89.3          & 52.8      & 61.4            & 91.5 & 71.3    & 80.6     & 88.3          & 50.9 & 58.3\\
{}\cite{melas-kyriazi2022deep}   & Kyriazi \emph{et~al.}                                            & --          & 76.9       & --           & --          & 51.4          & --        & -- & 73.3         & --      & --    & 56.7 & --\\
{}\cite{yang2021dystab} & DyStaB$^\dagger$ & -- & -- & -- & -- & -- & -- & -- & -- & 88.1 & -- & -- & 73.9 \\
{}\cite{wang2022selfsupervised} & TokenCut    & -- & -- & -- & 90.3 & 57.6 & -- & 91.8 & 71.2 & -- & 88.0 & 53.3 & -- \\
{}\cite{shin2022unsupervised} & SelfMask    & -- & -- & -- & 92.3 & 62.6 & -- & 94.4 & 78.1 & -- & 90.1 & 58.2 & -- \\
\midrule
\multicolumn{2}{l}{ \textbf{Ours}}                                            & 93.5 & 64.6 & 80.9     & 91.5 & 49.2 & 65.6  & 88.5 & 56.1 & 74.3                 & 89.3 & 41.31 & 56.3  \\
\bottomrule
\end{tabular}
\end{center}
\caption{\textbf{Unsupervised object segmentation} benchmark CUB and three saliency detection benchmarks: DUTS, ECSSD, and DUT-OMRON (\textit{OMRON}). $\dagger$ DyStaB uses CRF post-processing, supervised pre-training, and self-training on each dataset. (SoTA only table - please see the supplement for a complete version of this table including many older methods.) 
}\label{table:unsup_semseg_benchmark}
\end{table}
\subsection{Unsupervised Image Segmentation}

While the main aim of our work is object segmentation in videos, we also assess the image segmentation performance on common image segmentation and saliency benchmarks: CUB, DUTS, DUT-OMRON, and ECSSD.
For this experiment, we train our model on all three motion segmentation datasets (\DAVIS, \FBMS and \ST) jointly and apply the resulting network to the image segmentation benchmarks without any further fine-tuning. 
In \cref{table:unsup_semseg_benchmark}, we report the performance of our method and compare to the current state of the art.
It is worth noting that most prior work (except \cite{melas-kyriazi2022finding,melas-kyriazi2022deep,wang2022selfsupervised}) relies on dataset-specific training, self-training, post-processing or supervised pre-training to achieve image segmentation.

Finally, we evaluate the model qualitatively in \cref{fig:unsup_seg} on all four benchmarks.
We observe our model works well on a diverse set of classes, such as buildings, certain animals and plants, even though they were not part of the foreground (moving) objects in the training data.

\section{Conclusions}\label{s:conclusions}

We have proposed a simple approach to exploit the synergies between motion in videos and objectness for segmenting visual objects without supervision.
The key idea is using motion anticipation as a learning signal: we train
an image segmentation network to predict regions that likely contain simple optical flow patterns, 
as these have a high chance to correspond to objects.
We find that the complexity of the motion model is important to model complicated flow patters that can arise even for rigid objects. 
Our results show that this approach achieves state-of-the-art performance in video segmentation benchmarks.
Future work could thus consider extensions to more sophisticated motion models, accounting for the 3D shape of objects, and to separate multiple objects.

\newpage
\paragraph{Acknowledgments}
S.~C. is supported by a scholarship sponsored by Facebook.
L.~K. is funded by EPSRC Centre for Doctoral Training in Autonomous Intelligent Machines and Systems EP/S024050/1. 
I.~L. and A.~V. are supported by the European Research Council (ERC) grant IDIU-638009.
I.~L. is also supported by EPSRC grant VisualAI EP/T028572/1.
C.~R. is supported by Innovate UK (project 71653) on behalf of UK Research and Innovation (UKRI).

\bibliography{references}
\newpage
\appendix
\noindent
\section*{Supplementary Material}
In this supplementary material, we provide further details on our training parameters in~\Cref{sup:sec1}. \Cref{sup:sec2} contains the closed form solution of the fitting of the flow model $\theta$. Expanded experiments and ablations are found in \Cref{sup:sec3}. Finally, more qualitative results are presented in \cref{sup:sec4}. See the project page, \url{https://www.robots.ox.ac.uk/~vgg/research/gwm}, for additional visualizations, code and models.

\section{Experimental Setup}\label{sup:sec1}
\paragraph{Network.} We use MaskFormer~\cite{cheng2021maskformer} as our segmentation network\footnote{Implementation from  \url{https://github.com/facebookresearch/MaskFormer}.}, and use only the segmentation head. As MaskFormer predicts masks at 4 times lower resolution than input, we modify the \texttt{PixelDecoder} by appending [\textit{Conv}(3), \textit{UpsampleNN(2)}, \textit{Conv}(1)]$\times 2$ to its output layers to bring the masks back up to the input resolution.

For the backbone and appearance features $V$, we leverage a ViT-8 transformer, 
pre-trained on ImageNet~\cite{russakovsky2015imagenet} in a self-supervised manner using DINO~\cite{Caron_2021_ICCV}
to avoid any external sources of supervision. For the hierarchical backbone features to decoder we use the key feature outputs from layers 6, 8, 10, 12.

The input RGB images are interpolated (bi-cubic) to $128\times224$ resolution for input to the network. We interpolate (nearest neighbor) the optical flow to $480\times854$ for the loss.
Output segmentation logits are up-sampled using bi-linear interpolation to the flow resolution for training and again to annotation resolution for evaluation.

\paragraph{Training Hyperparameters.}\label{sup:hparam}
The networks are optimised using AdamW~\cite{loshchilov2018decoupled}, 
with learning rate of $1.5\times10^{-4}$, a schedule of linear warm-up from $1.0\times10^{-6}$ to $1.5\times10^{-4}$ over 1.5k iteration and polynomial decay afterwards. We use batch size of 8 and train for 15k iterations. We additionally employ gradient clipping when the 2-norm exceeds 0.01 for stability. The loss multiplier is $0.03$. 

\paragraph{UNet.} For experiments using U-Net\footnote{Implementation from \url{https://github.com/milesial/Pytorch-UNet}.}, we use the standard 4-layer version. The batch-size is increased to 16 and learning rate to $7.0\times10^{-4}$. We also clip the gradients only when 2-norm exceeds 5.0. All other settings, including optimizer and learning rate schedules, are kept the same. U-Net is not pre-trained and trained from scratch.  

\paragraph{Optical Flow.}
Our method derives its learning signal from optical flow estimated using off-the-shelf frozen networks.
We estimate optical flow for all frames on \DAVIS, \ST, and \FBMS following the practice of MotionGrouping~\cite{yang2021self-supervised}.
We employ RAFT~\cite{teed2020raft} (supervised) using the original resolution for our main experiments, and gaps between frames of $\{-2, -1, 1, 2\}$ for \DAVIS and \ST, and $\{-6, -3, 3, 6\}$ on \FBMS{}\@. When multiple flows are associated with a single frame (multiple gaps), we sample one at random for each iteration.

\section{Quadratic Flow Model: Closed Form Solution} \label{sup:sec2}

\newcommand{\USigma}{\Lambda}

Consider one of $K$ regions $m$ and define $w_u \propto P(m_u=k|I,\Phi)$ the posterior probability for that region, normalized so that $\sum_{u\in\Omega} w_u = 1$ (the scaling factor does not matter for the purpose of finding the minimizer).
We can obtain the minimizer $(A^*,b^*)$ and minimum of the energy
\begin{equation}\label{e:energy}
E(A,b)
=
\sum_{u \in \Omega} w_u \|F_u - Au-b\|^2
\end{equation}
as follows.
Defining
$$
\bar u := \begin{bmatrix} u \\ 1 \end{bmatrix},
~~~~
M := \begin{bmatrix} A & ~~b \end{bmatrix} \in \mathbb{R}^{2\times 6}
$$
allows rewriting the energy as
$$
E(M)
=
\sum_{u \in \Omega} w_u \|F_u - M \bar u\|^2
=
\operatorname{tr}\left(
\USigma_{FF} 
- M \USigma_{\bar \Omega F} 
- \USigma_{F \bar \Omega} M^\top 
+ M \USigma_{\bar \Omega \bar \Omega} M^\top
\right)  ,
$$
where
$$
\USigma_{FF} = \sum_{u \in \Omega} w_u F_uF_u^\top,~~
\USigma_{F\bar \Omega} = \sum_{u \in \Omega} w_u F_u\bar u^\top,~~
\USigma_{\bar \Omega F} = \USigma_{F\bar \Omega}^\top,~~
\USigma_{\bar \Omega \bar \Omega} = \sum_{u \in \Omega} w_u \bar u \bar u^\top.
$$
are the (uncentered) second moment matrices of the flow $F_u$ and homogeneous coordinate vectors $\bar u$.
By inspection of the trace term, the gradient of the energy is given by:
$$
\frac{dE(M)}{dM}
=
2\left(
\USigma_{F \bar\Omega}
-
M \USigma_{\bar\Omega \bar\Omega}
\right)
$$
Hence, the optimal regression matrix $M^*$ and corresponding energy value are
$$
M^* = \USigma_{F \bar \Omega} \USigma_{\bar \Omega \bar \Omega}^{-1},~~~~
E(M^*) = \operatorname{tr}\left(
\USigma_{FF} 
- M^* \USigma_{\Omega \bar F}
\right).
$$

Somewhat more intuitive results can be obtained by centering the moments and resolving for $A$ and $b$ instead of $M$.
Specifically, define:
$$
\mu_\Omega := \sum_{u\in\Omega} w_u u,~~~
\mu_F := \sum_{u\in\Omega} w_u F_u.
$$
The covariance matrices of the vectors are:
\begin{multline*}
\Sigma_{FF} = \sum_{u \in \Omega} w_u (F_u - \mu_F)(F_u-\mu_F)^\top,~~~
\Sigma_{F \Omega} = \sum_{u \in \Omega} w_u (F_u-\mu_F)(u-\mu_\Omega)^\top,\\
\Sigma_{\Omega F} = \USigma_{F\Omega}^\top,~~~
\Sigma_{\Omega \Omega} = \sum_{u \in \Omega} w_u (u - \mu_\Omega) (u - \mu_\Omega)^\top.
\end{multline*}
It is easy to check that
\begin{multline*}
\USigma_{FF} = \Sigma_{FF} + \mu_F\mu_F^\top, ~~~
\USigma_{F\bar \Omega} = \begin{bmatrix}
\Sigma_{F\Omega} + \mu_F\mu_\Omega^\top &~~ \mu_F
\end{bmatrix}, ~~~
\USigma_{\bar\Omega \bar \Omega} = \begin{bmatrix}
\Sigma_{\Omega\Omega} + \mu_\Omega\mu_\Omega^\top &~~ \mu_\Omega \\
\mu_\Omega^\top & ~~1
\end{bmatrix}.
\end{multline*}
From this:
\begin{multline*}
M^* 
= \USigma_{F \bar \Omega} \USigma_{\bar \Omega \bar \Omega}^{-1} =
\begin{bmatrix}
\Sigma_{F\Omega} + \mu_{F}\mu_{\Omega}^\top & ~~\mu_{F}
\end{bmatrix}
\begin{bmatrix}
\Sigma_{\Omega\Omega} + \mu_{\Omega}\mu_{\Omega}^\top & ~~\mu_{\Omega}\\
\mu_{\Omega}^\top & ~~1 \\
\end{bmatrix}^{-1}
\\
=
\begin{bmatrix}
\Sigma_{F\Omega} + \mu_{F}\mu_{\Omega}^\top & ~~\mu_{F}
\end{bmatrix}
\begin{bmatrix}
\Sigma_{\Omega\Omega}^{-1} & ~~-\Sigma_{\Omega\Omega}^{-1} \mu_{\Omega}\\
- \mu_{\Omega}^\top\Sigma_{\Omega\Omega}^{-1}  & ~~1 + \mu_{\Omega}^\top \Sigma_{\Omega\Omega}^{-1}  \mu_{\Omega}\\
\end{bmatrix}
\\
=
\begin{bmatrix}
\Sigma_{F\Omega} \Sigma_{\Omega\Omega}^{-1} & 
~~\mu_F  - \Sigma_{F\Omega}\Sigma_{\Omega\Omega}^{-1}\ \mu_\Omega
\end{bmatrix}
= \begin{bmatrix} A^* & ~~b^*\end{bmatrix}.
\end{multline*}
Hence, the optimal regression coefficients and energy value are also given by:
$$
A^* = \Sigma_{F\Omega} \Sigma_{\Omega\Omega}^{-1}, ~~~
b^* = \mu_F - A^* \mu_\Omega.%
$$

\section{Further Experiments} \label{sup:sec3}
\subsection{Generalization in Unsupervised Video Segmentation} 
\label{sup:video_gen}
\begin{table}
\begin{center}
\footnotesize
\begin{tabular}{rlccc}
\toprule
& Model & Flow & \textbf{\DAVIS}~($\mathcal{J}$$\uparrow$) & \textbf{\FBMS}~($\mathcal{J}$$\uparrow$) \\
\midrule
\cite{liu2021emergence} & AMD (100 steps) & \xmark & 57.8 & 47.5 \\
& \textbf{Ours} (Zero shot) & ARFlow & 62.5 & 65.4 \\
& \textbf{Ours} (20 steps) & ARFlow & 65.2 & 67.6 \\
\midrule
\cite{meunier2022em-driven} & EM & RAFT & 69.3 & 57.8 \\
& \textbf{Ours} (Zero shot) & RAFT & 66.8 & 73.2 \\
& \textbf{Ours} (20 steps) & RAFT & 76.3 & 77.1 \\
\bottomrule
\end{tabular}
\end{center}   
\caption{\textbf{Generalization performance on unseen \emph{videos}.} 
    Few unsupervised methods operate in this setting. AMD trains on YT-VOS, followed by 100 test-time adaptation steps, while EM trains on FlyingThings3D using flow as input. We use (fully unsupervised) ARFlow for fair comparison with AMD. Our method shows better performance after observing motion. (Test-time adaptation uses the training loss. No GT is involved at any point.)}
\label{tab:vid_gen}
\end{table}
We also test our model in a video \emph{generalization} setting. In contrast to the protocol of~\cite{yang-loquercio2019unsupervised,yang2021self-supervised}, where evaluation set is observed together with training to infer masks jointly\footnote{Note, no annotations are observed at any point.}, here we train only on frames from the training set.
We report performance on unseen videos. In this case, our method independently segments a collection of frames from a new video, with no way to incorporate motion information.

To ``observe'' motion on \emph{unseen} inputs, we also report results after taking 20 test-time adaptation steps (using our unsupervised loss) for each evaluation sequence in isolation (c.f. AMD \cite{liu2021emergence} takes 100 test-time adaptations steps). That is after training, 
we follow our training setup (optimizer, rate, batch size) and feed frames from the evaluation video and corresponding optical flow, calculate loss and take gradient steps. 
Despite other methods using much larger training sets, our approach shows better performance (\cref{tab:vid_gen}).

\subsection{Ablation Studies}\label{sup:ablation}

\paragraph{Pretraining.}\label{sup:pretraining}

\begin{table}[t]
\begin{center}
\footnotesize
\begin{tabular}{lccccc}
\toprule
Backbone & Backbone    &      & \textbf{\DAVIS} & \textbf{\ST} & \textbf{\FBMS} \\
model & pretraining & Sup. & $\mathcal{J}$$\uparrow$ & $\mathcal{J}$$\uparrow$ & $\mathcal{J}$$\uparrow$ \\  
\midrule
ViT-8 & ImageNet DINO & \xmark & 79.5 & 78.3 & 77.4 \\ 
UNet & None & \xmark & {78.3}     & {76.8}        & {72.0} \\
\midrule
SWIN-tiny & ImageNet MOBY  & \xmark & 78.3 & 77.4 & 74.6 \\
SWIN-tiny & ImageNet CLS  & \cmark & 78.9 & 77.7 & 75.5 \\
SWIN-tiny & None & \xmark & 78.3 & 75.2 & 68.8 \\
Resnet-50 & ImageNet CLS  & \cmark & 77.5 & 75.8 & 72.9 \\

\bottomrule
\end{tabular}
\end{center}
\caption{\textbf{Effect of Pretraining/Backbone.} Our method with MaskFormer benefits from pretraining, with slight improvement offered by supervised (\textit{CLS}) over unsupervised (\textit{MOBY}) pretraining (usng SWIN transformer). Comparable results can be obtained with training from scratch. Best results are obtained using DINO features.}
\label{tab:weights_ablation}
\end{table}
\begin{table}[t]
\begin{center}
\footnotesize
\begin{tabular}{lccccc}
\toprule
 &     &      & \textbf{\DAVIS} & \textbf{\ST} & \textbf{\FBMS} \\
Model & $K$ &  Merge & $\mathcal{J}$$\uparrow$ & $\mathcal{J}$$\uparrow$ & $\mathcal{J}$$\uparrow$ \\  
\midrule
Ours & $K=4$ & \cmark & \textbf{79.5} & \textbf{78.3} & \textbf{77.4} \\
\midrule
Spectral clustering & $K=2$ & \xmark & 15.79 & 14.89 & 27.45 \\
\midrule
K-Means & $K=4$  & \cmark & 41.79 & 34.84 & 48.80 \\
K-Means & $K=2$  & \xmark & 20.24 & 21.14 & 38.25 \\

\bottomrule
\end{tabular}
\end{center}
\caption{\textbf{Feature Clustering without Motion.} We experiment with offline clustering of DINO features to assess the importance of our motion-based formulation. Simply clustering DINO features using K-Means or spectral clustering \cite{melas-kyriazi2022deep} into 2 clusters performs worse. Over-clustering and merging using our cluster-merging approach performs better but still fails to reach our performance.}
\label{tab:dino_ablation}
\end{table}

Compared to recent methods for video segmentation~\cite{yang2021self-supervised,meunier2022em-driven}, one of the benefits of our formulation is that we can leverage unsupervised pretraining for the segmentation network (\eg, for the ViT backbone of MarkFormer).
This enables our method to be trained in only 15k iterations.
Here, we investigate the importance of the backbone. To this end we replace ViT with Swin-tiny pretrained using MOBY (self-supervised) in \cref{tab:weights_ablation}. The performance differences are small.

Additionally, we investigate the effect of other pretraining strategies on the performance.
Switching to a model pretrained on ImageNet with image-level supervision (\ie a classification task) 
only slightly improves performance showing that the method does not need to rely on supervised pre-training. %
Finally, we train the model using same settings for 20k iterations from scratch, without any pre-training.
This results in comparable performance on \DAVIS but reduced performance on the smaller datasets.
Comparing backbones without pre-training, UNet gives better results than SWIN-tiny, likely due to smaller networks being easier to train on small datasets.

\paragraph{Feature Clustering without Motion.}
To demonstrate the potential of using motion for discovering objects, in \cref{tab:dino_ablation}, we compare to additional baselines that only rely on clustering visual features. Spectral feature clustering with $K=2$ (based on~\cite{melas-kyriazi2022deep}), on the same visual features we use to merge segments (\ie, DINO) after over-clustering, shows (somewhat unsurprisingly) that learning from motion is important for motion segmentation. Similarly, K-means ($K=2$) on the same features also falls behind our method. Yet, we show that K-means also benefits from over-clustering ($K=4$) and then merging. 

\paragraph{Flow Estimation.}\label{sup:flow}

\begin{table}[t]
\parbox{.48\linewidth}{
    \begin{center}
    \footnotesize
    \begin{tabular}{rlccc}
    \toprule
    \multicolumn{2}{l}{Opt. Flow} & Sup. & \textbf{\DAVIS}~($\mathcal{J}$$\uparrow$) \\ 
    \midrule
    \cite{liu2020learning} & ARFlow    & \xmark & 66.9 \\
    \cite{sun2018pwcnet}   & PWCNet    & \cmark & 74.9 \\
    \cite{teed2020raft}    & RAFT      & \cmark & 79.5 \\
    \bottomrule
    \end{tabular}%
    \end{center}
    \caption{\textbf{Choice of Optical Flow Method.} Measuring the influence of the method to extract optical flow.}
    \label{tab:flow_ablation}
}
\hfill
\parbox{.48\linewidth}{
    \begin{center}
    \footnotesize
    \begin{tabular}{rlccc}
    \toprule
    \multicolumn{2}{l}{Method} & \textbf{\DAVIS}~($\mathcal{J}$$\uparrow$) \\ 
    \midrule
    \cite{yang2021self-supervised} & MG & 53.2 \\
    \cite{liu2021emergence} & AMD & 57.8 \\
    \midrule
     & {\bf Ours}     & {\bf 66.9} \\
    \bottomrule
    \end{tabular}%
    \end{center}
    \caption{\textbf{Fully Unsupervised Video Object Segmentation.} Comparison to the state of the art in unsupervised VOS without reliance on \textit{any} supervision
    }
    \label{tab:unsup_flow_results}
}
\end{table}

\begin{table}[t]
\footnotesize
\begin{center}
\setlength{\tabcolsep}{4.5pt}
\renewcommand*{\arraystretch}{0.95}
\begin{tabular}{@{}r@{\hspace{2pt}}lccc@{\hspace{10pt}}ccc@{\hspace{10pt}}ccc@{\hspace{10pt}}ccc} 
\toprule
& \multicolumn{1}{c}{}  & \multicolumn{3}{c}{\hspace{-20pt}\textbf{CUB}} & \multicolumn{3}{c}{\hspace{-20pt}\textbf{DUTS}} & \multicolumn{3}{c}{\hspace{-20pt}\textbf{ECSSD}}& \multicolumn{3}{c}{\hspace{-3pt}\textbf{OMRON}} \\ 
\cmidrule(r{10pt}){3-5}
\cmidrule(r{10pt}){6-8}
\cmidrule(r{10pt}){9-11}
\cmidrule{12-14}
                                    &                                                         & Acc           & $\mathcal{J}\uparrow$ & $max F_{\beta}\uparrow$  & Acc           & $\mathcal{J}\uparrow$  & $F_{\beta}\uparrow$  & Acc           & $\mathcal{J}\uparrow$  & $F_{\beta}\uparrow$   &    Acc           &$\mathcal{J}\uparrow$ & $F_{\beta}\uparrow$  \\
\midrule
{}\cite{xia2017wnet}                & WNet$^{\dagger}$                                        & --            & 24.8
& -- & -- & -- & -- & -- & -- & -- & -- & -- & -- \\
{}\cite{ji19invariant}              & IIC-seg                                                 & --            & 36.5
& -- & -- & -- & -- & -- & -- & -- & -- & -- & -- \\
{}\cite{bielski2019emergence}       & PertGAN                                                 & --            & 38.0
& -- & -- & -- & -- & -- & -- & -- & -- & -- & -- \\
{}\cite{chen2019unsupervised}       & ReDO                                                    & 84.5          & 42.6
& -- & -- & -- & -- & -- & -- & -- & -- & -- & -- \\
{}\cite{kanezaki2018unsupervised}   & UISB                                                    & --            & 44.2
& -- & -- & -- & -- & -- & -- & -- & -- & -- & -- \\
{}\cite{benny2020onegan}            & OneGAN                                                  & --            & 55.5 
& -- & -- & -- & -- & -- & -- & -- & -- & -- & -- \\
{}\cite{yu2021unsupervised}        & DRC & --            & 56.4
& -- & -- & -- & -- & -- & -- & -- & -- & -- & -- \\
{}\cite{he2021ganseg}               & GANSeg       & --            & 62.9 & -- & -- & -- & -- & -- & -- & -- & -- & -- & -- \\       
{}\cite{voynov2021object}         & Voynov \emph{et~al.}                                                 & 94.0          & 71.0    & 80.7              & 88.1          & 51.1     & 60.0             & 90.6          & 68.4      & 79.0            & 86.0          & 46.4 & 53.3\\
{}\cite{liu2021emergence} & AMD & -- & -- & -- & -- & -- & 60.2 & -- & -- & -- & -- & -- & -- \\
{}\cite{melas-kyriazi2022finding}   & Kyriazi \emph{et~al.}                                            & 92.1          & 66.4      & 78.3            & 89.3          & 52.8      & 61.4            & 91.5 & 71.3    & 80.6     & 88.3          & 50.9 & 58.3\\
{}\cite{melas-kyriazi2022deep}   & Kyriazi \emph{et~al.}                                            & --          & 76.9       & --           & --          & 51.4          & --        & -- & 73.3         & --          & 56.7 & --\\
{}\cite{yang2021dystab} & DyStaB$^\dagger$ & -- & -- & -- & -- & -- & -- & -- & -- & 88.1 & -- & -- & 73.9 \\
{}\cite{wang2022selfsupervised} & TokenCut    & -- & -- & -- & 90.3 & 57.6 & -- & 91.8 & 71.2 & -- & 88.0 & 53.3 & -- \\
{}\cite{shin2022unsupervised} & SelfMask    & -- & -- & -- & 92.3 & 62.6 & -- & 94.4 & 78.1 & -- & 90.1 & 58.2 & -- \\
\midrule
\multicolumn{2}{l}{ \textbf{Ours}}                                            & 93.5 & 64.6 & 80.9     & 91.5 & 49.2 & 65.6  & 88.5 & 56.1 & 74.3                 & 89.3 & 41.31 & 56.3  \\
\bottomrule
\end{tabular}
\end{center}
\caption{\textbf{Expanded unsupervised object segmentation} benchmark CUB and three saliency detection benchmarks: DUTS, ECSSD, and DUT-OMRON (\textit{OMRON}). $\dagger$ DyStaB uses CRF post-processing, supervised pre-training, and self-training on each dataset.}
\label{table:unsup_semseg_expanded}
\end{table}

Finally, our method relies on optical flow estimated by frozen, off-the-shelf networks.
So far we have been using RAFT~\cite{teed2020raft}, as such optical flow network was adopted in our baselines.
In \cref{tab:flow_ablation}, we also consider PWCNet~\cite{sun2018pwcnet} and fully-unsupervised ARFlow~\cite{liu2020learning}.
We observe that the performance of the flow estimator has an impact on the final performance of our method.
Finally, we compare our \emph{fully} unsupervised model (which uses self-supervised pretraining and flow) to fully unsupervised state-of-the-art methods.
Appearance-Motion Decomposition (AMD)~\cite{liu2021emergence} works end-to-end and directly extracts motion features from pairs of images with a PWCNet-like architecture, while MotionGrouping (MG)~\cite{yang2021self-supervised} and our method use ARFlow~\cite{liu2020learning} for optical flow estimation.
In \cref{tab:unsup_flow_results} we show that our method achieves a significant improvement over previous approaches.

\section{Additional Results and Discussion}\label{sup:sec4}
We provide a further breakdown of our results in \cref{tab:supp_seq_davis,tab:supp_seq_stv2,tab:supp_seq_fbms}, reporting per sequence evaluation results on the video segmentation tasks.

\paragraph{Video object segmentation and egomotion.}
We note that some sequences have pronounced egomotion (\eg, camera shaking in \texttt{libby} of DAVIS or inside a moving car in \texttt{camel01} of FBMS). 
Our model performs well on these sequences, demonstrating that it can handle egomotion. 
When \emph{only} the camera is moving, the resulting optical flow would still highlight objects due to parallax.
This provides a learning signal, however, it would likely be
weaker for objects farther away from the camera.
As our method works on a per-frame basis and does not \emph{require} flow during inference, this should not have an impact at test time. 
However, fine-tuning on scenes with only egomotion (see \cref{sup:video_gen} for experiments investigating test-time adaptation)   and only small or far away objects, might lead to the model learning to ignore them.

\begin{figure}[t]
\centering
\begin{tabular}{p{4pt}c}
    & \textbf{CUB} \\[-1pt]
    \rotatebox[origin=l]{90}{\scriptsize \hspace{10pt} Ours \hspace{28pt} GT  \hspace{30pt} Image \hspace{25pt} Ours \hspace{30pt} GT  \hspace{26pt} Image \hspace{38pt} Ours \hspace{30pt} GT  \hspace{30pt} Image} 
     & \includegraphics[page=1,width=.55\textheight,valign=l]{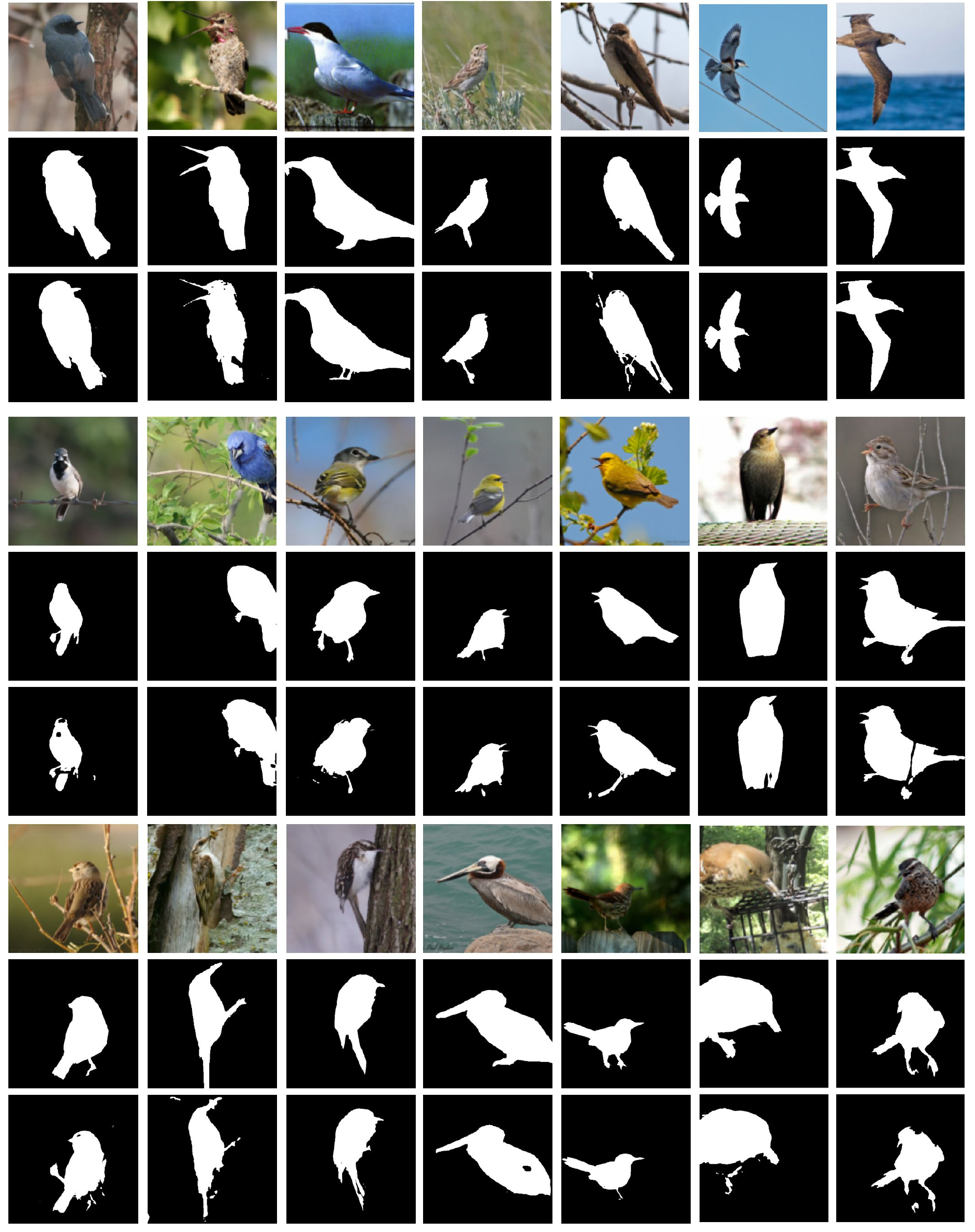} 
\end{tabular}
\caption{\textbf{Qualitative Comparison on CUB}. We train our model on a combined dataset of \DAVIS,~\FBMS~and~\ST. Our method can extract birds in different environments and poses. Our model can segment different species of birds}
\label{fig:unsup_seg_supp1}
\end{figure}

\begin{figure}[t]
\centering
\begin{tabular}{p{4pt}c}
    & \textbf{DUT-OMRON} \\[-1pt]
    \rotatebox[origin=l]{90}{\scriptsize  \hspace{5pt} Ours \hspace{20pt} GT  \hspace{20pt} Image \hspace{26pt} Ours \hspace{35pt} GT  \hspace{37pt} Image \hspace{30pt} Ours \hspace{31pt} GT  \hspace{30pt} Image} 
     & \includegraphics[page=2,width=.55\textheight,valign=l]{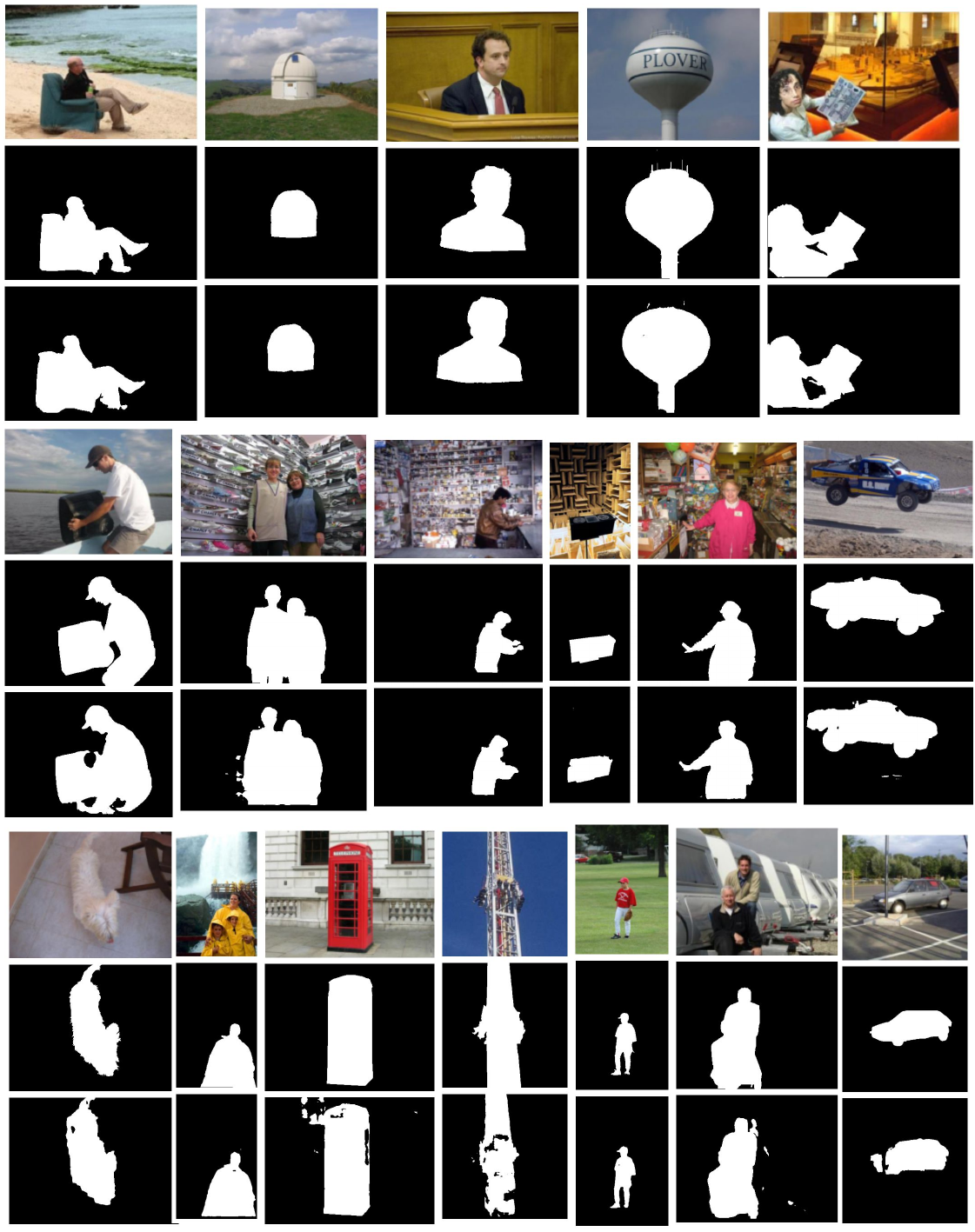} 
\end{tabular}
\caption{\textbf{Qualitative Comparison on DUT-OMRON}. We train our model on a combined dataset of \DAVIS,~\FBMS~and~\ST. Our model can segment both stationary and non-stationary objects and is robust enough to work on a wide range of classes }
\label{fig:unsup_seg_supp2}
\end{figure}

\begin{figure}[t]
\centering
\begin{tabular}{p{4pt}c}
    & \textbf{DUTS} \\[-1pt]
    \rotatebox[origin=l]{90}{\scriptsize  \hspace{15pt} Ours \hspace{22pt} GT  \hspace{35pt} Image \hspace{26pt} Ours \hspace{35pt} GT  \hspace{24pt} Image \hspace{25pt} Ours \hspace{25pt} GT  \hspace{30pt} Image} 
     & \includegraphics[page=3,width=.55\textheight,valign=l]{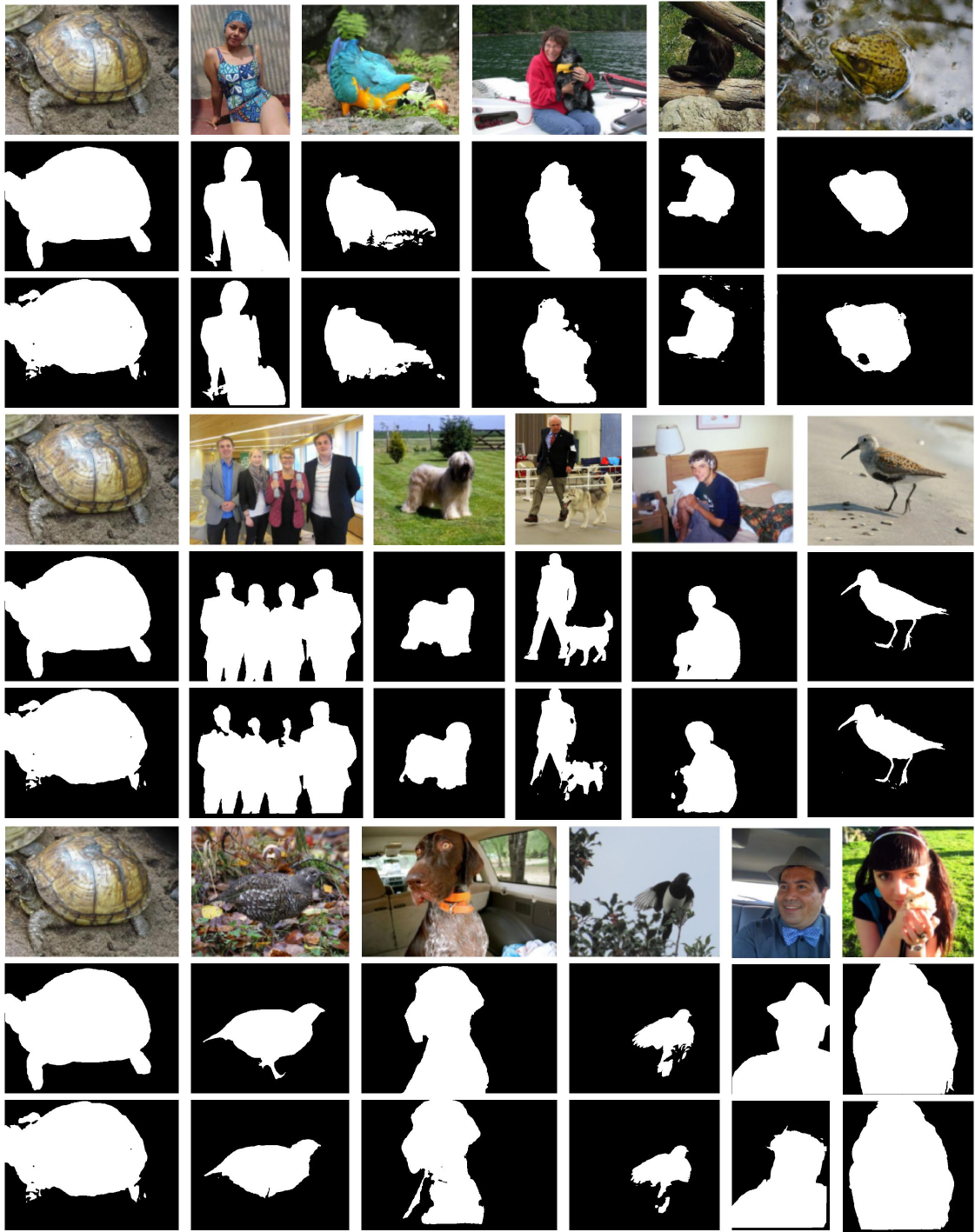} 
\end{tabular}
\caption{\textbf{Qualitative Comparison on DUTS}. We train our model on a combined dataset of \DAVIS,~\FBMS~and~\ST. We can segment a wide array of classes. Our model performs well on scenes where multiple objects are in the foreground}
\label{fig:unsup_seg_supp3}
\end{figure}

\begin{figure}[t]
\centering
\begin{tabular}{p{4pt}c}
    & \textbf{ECSSD} \\[-1pt]
    \rotatebox[origin=l]{90}{\scriptsize  \hspace{15pt} Ours \hspace{35pt} GT  \hspace{38pt} Image \hspace{35pt} Ours \hspace{25pt} GT  \hspace{27pt} Image \hspace{25pt} Ours \hspace{32pt} GT  \hspace{35pt} Image} 
     & \includegraphics[page=4,width=.6\textheight,valign=l]{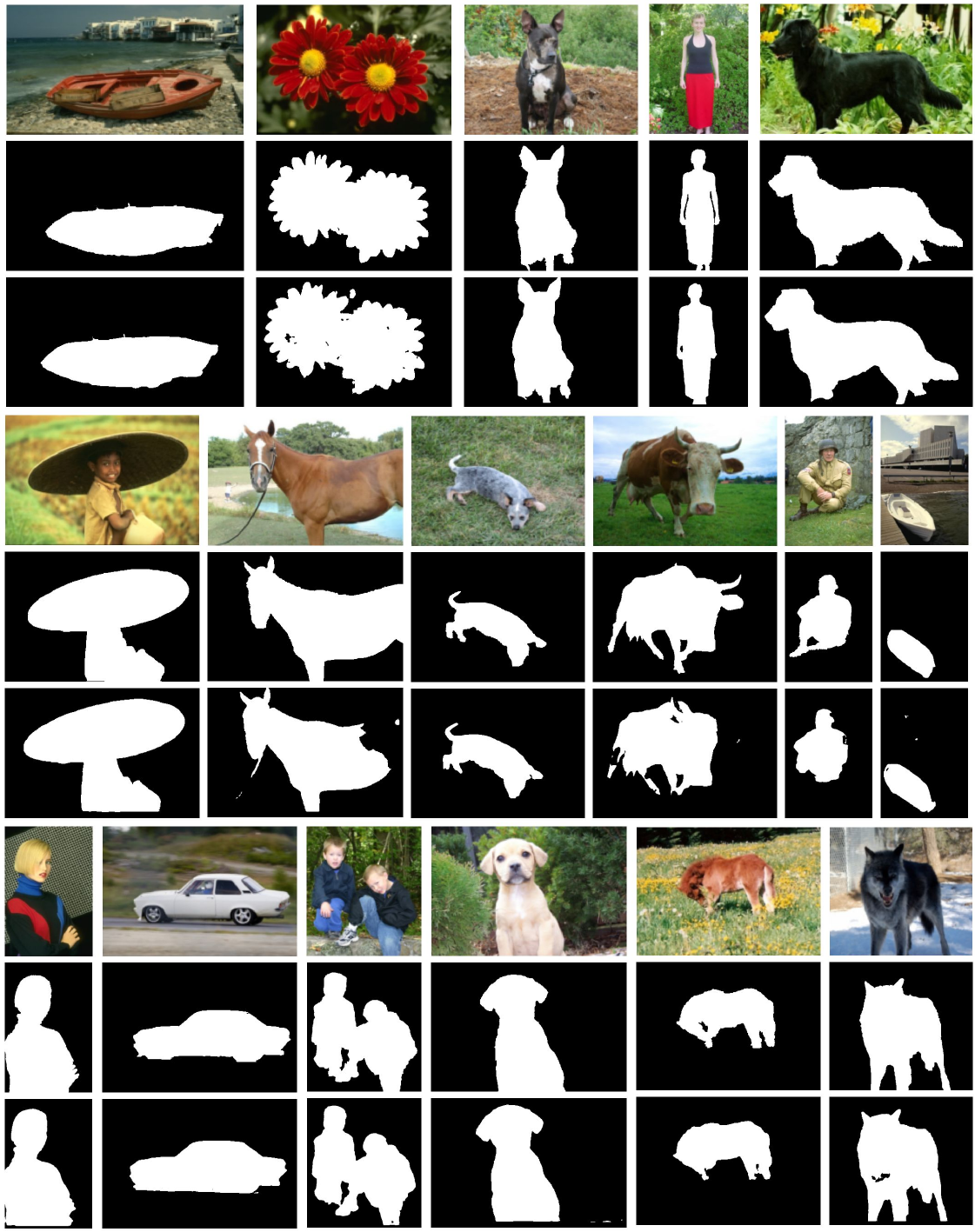} 
\end{tabular}
\caption{\textbf{Qualitative Comparison on ECSSD}. We train our model on a combined dataset of \DAVIS,~\FBMS~and~\ST. Our model can segment objects from different classes in complex poses}
\label{fig:unsup_seg_supp4}
\end{figure}

\begin{figure}[t]
\centering
\begin{tabular}{p{4pt}c}

    \rotatebox[origin=l]{90}{\scriptsize  \hspace{5pt} Ours \hspace{15pt} GT  \hspace{14pt} Image \hspace{22pt} Ours \hspace{18pt} GT  \hspace{12pt} Image \hspace{22pt} Ours \hspace{16pt} GT  \hspace{11pt} Image \hspace{25pt} Ours \hspace{22pt} GT  \hspace{25pt} Image} 
     & \includegraphics[page=5,width=.6\textheight,valign=l]{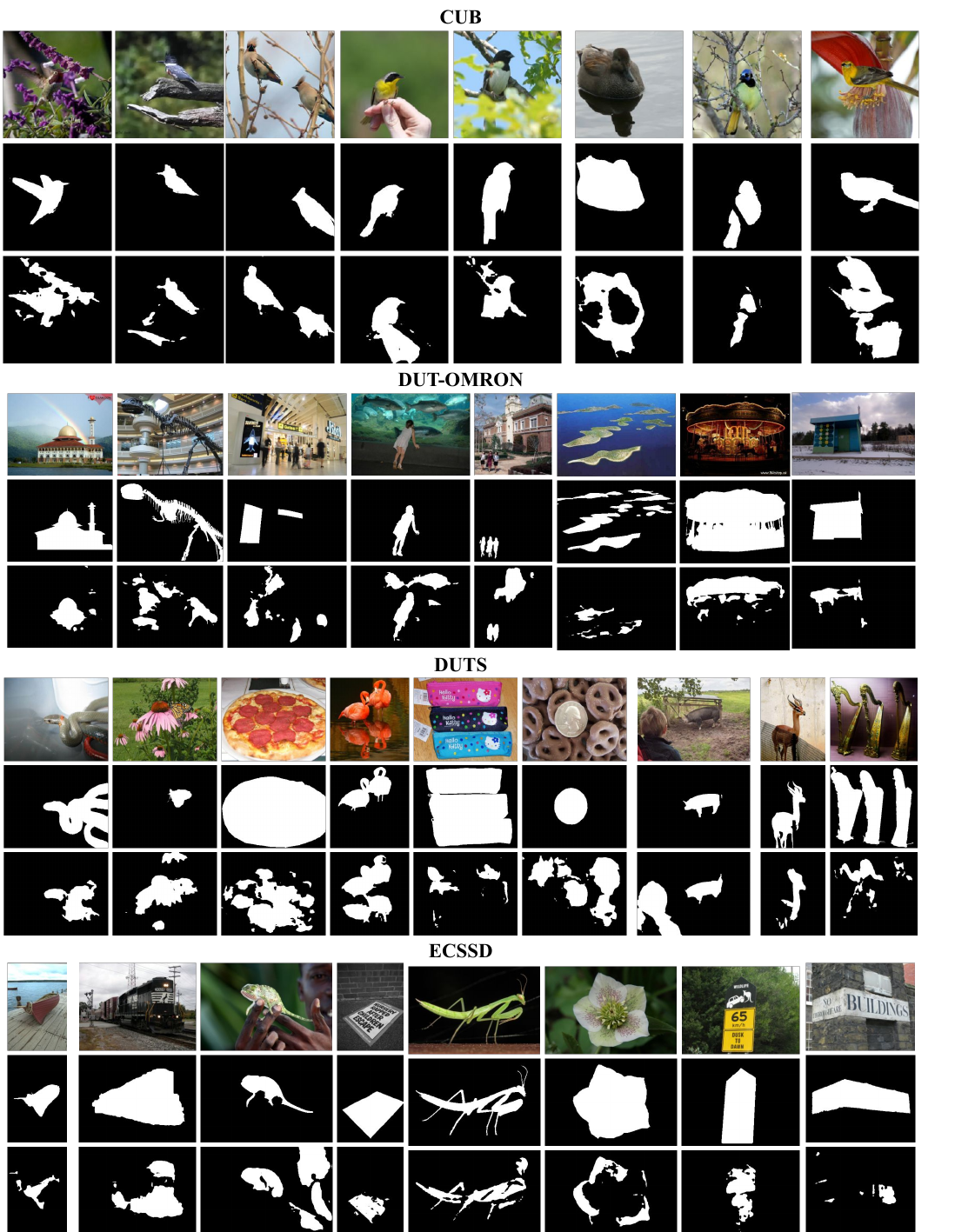} 
\end{tabular}
\caption{\textbf{Qualitative Comparison of Failure Cases}. We train our model on a combined dataset of \DAVIS,~\FBMS~and~\ST. Our method can extract salient object in various environments. The model has difficulty where the foreground object is ambiguous\,---\,when there are multiple prominent objects but only few are annotated as salient object. The model also has issues with predicting the object boundaries well for some instances  }
\label{fig:unsup_seg_supp5}
\end{figure}

\paragraph{Image segmentation.}
For unsupervised image segmentation, we show some additional qualitative results for CUB in \cref{fig:unsup_seg_supp1}, DUT-OMRON in \cref{fig:unsup_seg_supp2}, DUTS in \cref{fig:unsup_seg_supp3}, and ECSSD in \cref{fig:unsup_seg_supp4}. Our model, trained on a combined dataset of \DAVIS,~\FBMS~and~\ST, is robust enough to handle a wide array of classes from the above datasets in varying context. Our model can segment both stationary and non-stationary objects and works well when multiple objects are in the foreground.
In \cref{fig:unsup_seg_supp5}, we show a few failure cases for all datasets, where the model struggles mostly with ambiguous foreground objects and, in particular, with close-ups of stationary objects, \eg signs (ECSSD) and buildings (DUT-OMRON).
The model also has issues with boundaries for many objects, \ie the foreground objects are correctly identified but the model fails to fully segment them. 
For example, in DUTS, the snake in the first image has a well segmented head, however, the model does not segment its body accurately.

\begin{table}[t]
    \begin{center}
    \begin{tabular}{cccc@{\hspace{10pt}}ccc}
        \toprule
         & \multicolumn{3}{c}{\textit{w/o CRF}} & \multicolumn{3}{c}{\textit{w/ CRF}} \\
         Sequence & $\mathcal{J}$(M) & $\mathcal{J}$(R) & $\mathcal{J}$(D) & $\mathcal{J}$(M) & $\mathcal{J}$(R) & $\mathcal{J}$(D)  \\
         \midrule
    blackswan      &  67.0 &  100.0 &  -0.8 & 67.4 &  100.0 &    1.1 \\
    bmx-trees      &  58.2 &   76.9 &  19.9 & 59.8 &   76.9 &   17.5 \\
    breakdance     &  86.2 &  100.0 &   4.9 & 87.4 &  100.0 &    5.2 \\
      camel        &  89.4 &  100.0 &   5.7 & 90.6 &  100.0 &    5.5 \\
  car-roundabout   &  81.4 &   90.4 &  26.7 & 81.2 &   90.4 &   25.8 \\
    car-shadow     &  84.3 &  100.0 &   9.0 & 83.9 &  100.0 &    8.0 \\
       cows        &  90.4 &  100.0 &   3.4 & 91.3 &  100.0 &    3.2 \\
   dance-twirl     &  87.4 &  100.0 &  -7.1 & 88.8 &  100.0 &   -6.2 \\
       dog         &  92.9 &  100.0 &  -1.7 & 93.9 &  100.0 &   -1.6 \\
  drift-chicane    &  78.6 &   98.0 &   2.2 & 82.0 &  100.0 &    2.6 \\
  drift-straight   &  80.6 &  100.0 &   7.2 & 82.1 &  100.0 &    8.2 \\
       goat        &  78.6 &  100.0 &   1.7 & 75.8 &  100.0 &    4.5 \\
  horsejump-high   &  84.9 &  100.0 &   6.4 & 88.0 &  100.0 &    4.6 \\
    kite-surf      &  64.4 &   97.9 &   4.5 & 67.5 &   97.9 &    3.1 \\
      libby        &  82.9 &  100.0 &   8.6 & 84.5 &  100.0 &    8.6 \\
  motocross-jump   &  74.1 &   78.9 &   4.1 & 75.1 &   81.6 &    4.1 \\
paragliding-launch &  62.2 &   65.4 &  33.5 & 64.1 &   66.7 &   35.8 \\
     parkour       &  86.1 &  100.0 &  -4.5 & 88.1 &  100.0 &   -3.1 \\
  scooter-black    &  82.1 &   97.6 &  -4.3 & 82.1 &  100.0 &   -4.3 \\
     soapbox       &  79.2 &  100.0 &  -2.8 & 81.0 &  100.0 &   -0.4 \\
\midrule
     Average       &  79.5 &   95.3 &   5.8 & 80.7 &   95.7 &    6.1 \\
            \bottomrule
    \end{tabular}
    \end{center}
    \caption{\textbf{Result breakdown on DAVIS16 validation sequences.} (\textit{M}), (\textit{R}), and (\textit{D}) are mean, recall and decay of IoU, respectively}
    \label{tab:supp_seq_davis}
\end{table}

\begin{table}[t]
\parbox{.48\linewidth}{
    
    \begin{center}
    \begin{tabular}{ccc}
        \toprule
         & \textit{w/o CRF}  & \textit{w/ CRF} \\
         Sequence & $\mathcal{J}$(M) & $\mathcal{J}$(M)  \\
         \midrule
drift            & 86.1 & 86.5 \\
birdfall         & 67.8 & 57.1 \\
girl             & 84.5 & 86.3 \\
cheetah          & 57.0 & 50.8 \\
worm             & 83.7 & 84.0 \\
parachute        & 90.6 & 93.2 \\
monkeydog        & 22.9 & 22.6 \\
hummingbird      & 57.3 & 57.2 \\
soldier          & 77.4 & 77.4 \\
bmx              & 76.4 & 77.5 \\
frog             & 84.1 & 86.7 \\
penguin          & 77.7 & 76.8 \\
monkey           & 75.0 & 75.8 \\
bird of paradise & 92.3 & 94.0 \\
\midrule
Seq. Avg.          & 73.8 & 73.3 \\
Frame Avg.         & 78.3 & 78.9 \\
            \bottomrule
    \end{tabular}
    \end{center}
    \caption{Sequence breakdown on SegTrackv2 dataset. }
    \label{tab:supp_seq_stv2}
}
\hfill
\parbox{.48\linewidth}{
    \begin{center}
    \begin{tabular}{ccc}
        \toprule
         & \textit{w/o CRF}  & \textit{w/ CRF} \\
         Sequence & $\mathcal{J}$(M) & $\mathcal{J}$(M)  \\
         \midrule
camel01 & 86.8 & 91.0 \\
cars1 & 86.9 & 86.8 \\
cars10 & 64.6 & 64.8 \\
cars4 & 81.5 & 82.4 \\
cars5 & 81.6 & 82.1 \\
cats01 & 87.7 & 89.5 \\
cats03 & 69.4 & 63.2 \\
cats06 & 66.5 & 67.4 \\
dogs01 & 76.3 & 75.6 \\
dogs02 & 85.3 & 86.4 \\
farm01 & 90.8 & 90.5 \\
giraffes01 & 82.1 & 83.9 \\
goats01 & 79.9 & 83.7 \\
horses02 & 80.4 & 83.6 \\
horses04 & 59.8 & 60.5 \\
horses05 & 72.8 & 74.5 \\
lion01 & 75.1 & 75.0 \\
marple12 & 81.9 & 81.6 \\
marple2 & 84.4 & 85.9 \\
marple4 & 81.1 & 82.4 \\
marple6 & 95.1 & 95.1 \\
marple7 & 76.6 & 77.6 \\
marple9 & 95.4 & 96.3 \\
people03 & 90.1 & 91.0 \\
people1 & 85.3 & 87.2 \\
people2 & 88.1 & 89.7 \\
rabbits02 & 91.2 & 91.2 \\
rabbits03 & 81.5 & 84.4 \\
rabbits04 & 43.8 & 44.1 \\
tennis & 73.3 & 74.2 \\
\midrule
Seq. Avg.  & 79.8 & 80.7 \\
Frame Avg. & 77.4 & 78.4 \\
            \bottomrule
    \end{tabular}
    \end{center}
    \caption{Sequence breakdown on FBMS59 dataset}
    \label{tab:supp_seq_fbms}
}
\end{table}

\end{document}